\documentclass[review]{elsarticle}

\usepackage{amsmath}
\usepackage{times}
\usepackage{epsfig}
\usepackage{graphicx}
\usepackage{amsmath}
\usepackage{amssymb}

\usepackage{caption}
\usepackage{algorithm}
\usepackage{algorithmic}
\usepackage{float}
\usepackage{amsmath}
\usepackage{mathrsfs}

\journal{Journal of \LaTeX\ Templates}

\begin{document}

\begin{frontmatter}

\title{Structure propagation for zero-shot learning}


\author[mymainaddress]{Guangfeng Lin\corref{mycorrespondingauthor}}
\cortext[mycorrespondingauthor]{Corresponding author}
\ead{lgf78103@xaut.edu.cn}

\author[mymainaddress]{Yajun Chen}
\author[mymainaddress]{Fan Zhao}

\address[mymainaddress]{Information science department, Xi’an University of Technology,\\
 5 South Jinhua Road, Xi'an, Shaanxi Province 710048, PR China}

%
%

\begin{abstract}
The key of zero-shot learning (ZSL) is how to find the information transfer model for bridging the gap between images and semantic information (texts or attributes). Existing ZSL methods usually construct the compatibility function between images and class labels with the consideration of the relevance on the semantic classes (the manifold structure of semantic classes). However, the relationship of image classes (the manifold structure of image classes) is also very important for the compatibility model construction. It is difficult to capture the relationship among image classes due to unseen classes, so that the manifold structure of image classes often is ignored in ZSL. To complement each other between the manifold structure of image classes and that of semantic classes information, we propose structure propagation (SP) for improving the performance of ZSL for classification. SP can jointly consider the manifold structure of image classes and that of semantic classes for approximating to the intrinsic structure of object classes. Moreover, the SP can describe the constrain condition between the compatibility function and these manifold structures for balancing the influence of the structure propagation iteration. The SP solution provides not only unseen class labels but also the relationship of two manifold structures that encode the positive transfer in structure propagation. Experimental results demonstrate that SP can attain the promising results on the AwA, CUB, Dogs and SUN databases.
\end{abstract}


\end{frontmatter}

\section{Introduction}
Although deep learning \cite{Zhang2016Deep} depending on large-scale labeled data training has been widespreadly used for visual recognition, a daunting challenge still exists to recognize visual object "in the wild". In fact, in specific applications it is impossible to collect all class data for training deep model, so training and testing class sets are often disjoint. The main idea of ZSL is to handle this problem by exploit the transfer model via the redundant relevance of the semantic description. Furthermore, in ZSL, testing class images can be mapped into the semantic or label space by transfer model for recognizing objects of unseen classes, from which samples are not available or can not be collected in training sets. Many ZSL methods \cite{Larochelle2008Zero} \cite{Yu2010Attribute} \cite{Rohrbach2011} \cite{Lampert2014} \cite{Huang2015} \cite{Changpinyo2016} \cite{xian2017zero} attempt to model the interaction relationship on the cross-domain (e.g. text domain or image domain) via transfer model to classifying objects of unseen classes by the aid of the semantic description of unseen classes and seen classes, from which labeled samples can be used. For example, 'pig' is a unseen class in testing image sets, while 'zebra' is a seen class in training image sets. These classes both have the related semantic description(e.g. attribute is 'has tail'). Therefore, ZSL can construct a knowledge transfer model between 'zebra' and 'has tail' in training sets, and then, 'pig' can be mapped into the semantic or label space by this model for recognizing 'pig' in testing image sets.

To recognize unseen classes from seen classes, ZSL needs face to two challenges \cite{Changpinyo2016}. One is how to utilize the semantic information for constructing the relationship between unseen classes and seen classes, and other is how to find the compatibility among all kinds of information for obtaining the optimal discriminative characteristics on unseen classes.

To handle the first challenge, visual attributes \cite{farhadi2009describing} \cite{lampert2009learning} \cite{parikh2011relative} and text representations \cite{Frome2013DeViSE} \cite{mikolov2013efficient} \cite{Socher2013Zero} have been used for linking unseen and seen classes. These semantic information can not only be regarded as a middle bridge for associating visual images and class labels \cite{Socher2013Zero} \cite{jayaraman2014zero} \cite{lampert2014attribute} \cite{li2015semi} \cite{li2014attributes} \cite{norouzi2013zero} \cite{romera2015embarrassingly}, but also be transformed into new representations corresponding to the more suitable relation between unseen and seen classes by Canonical Correlation Analysis (CCA)\cite{fu2015transductive} or Sparse Coding (SC)\cite{yu2017transductive} \cite{zhang2015zero}. To address the second challenge, the classical method as baseline is the probability model for visual attributes predicting unseen class labels \cite{lampert2014attribute}. For implementing the discriminative classification, the recent methods have two tendencies. some methods have built the linear \cite{Frome2013DeViSE} \cite{7298911} \cite{7293699}, nonlinear \cite{Socher2013Zero} \cite{7780384} or hybrid \cite{norouzi2013zero} \cite{zhang2015zero} compatibility function between image domain and semantic domain (for example,text domain), others have further considered the manifold structure of semantic classes for constraining the compatibility function \cite{Changpinyo2016}. However, the manifold structure of image classes that can enhance the connection between unseen classes and seen classes is often neglected, because unseen classes make the manifold structure of image classes to be uncertain.

\begin{figure*}[ht]
  \begin{center}
\includegraphics[width=1\linewidth]{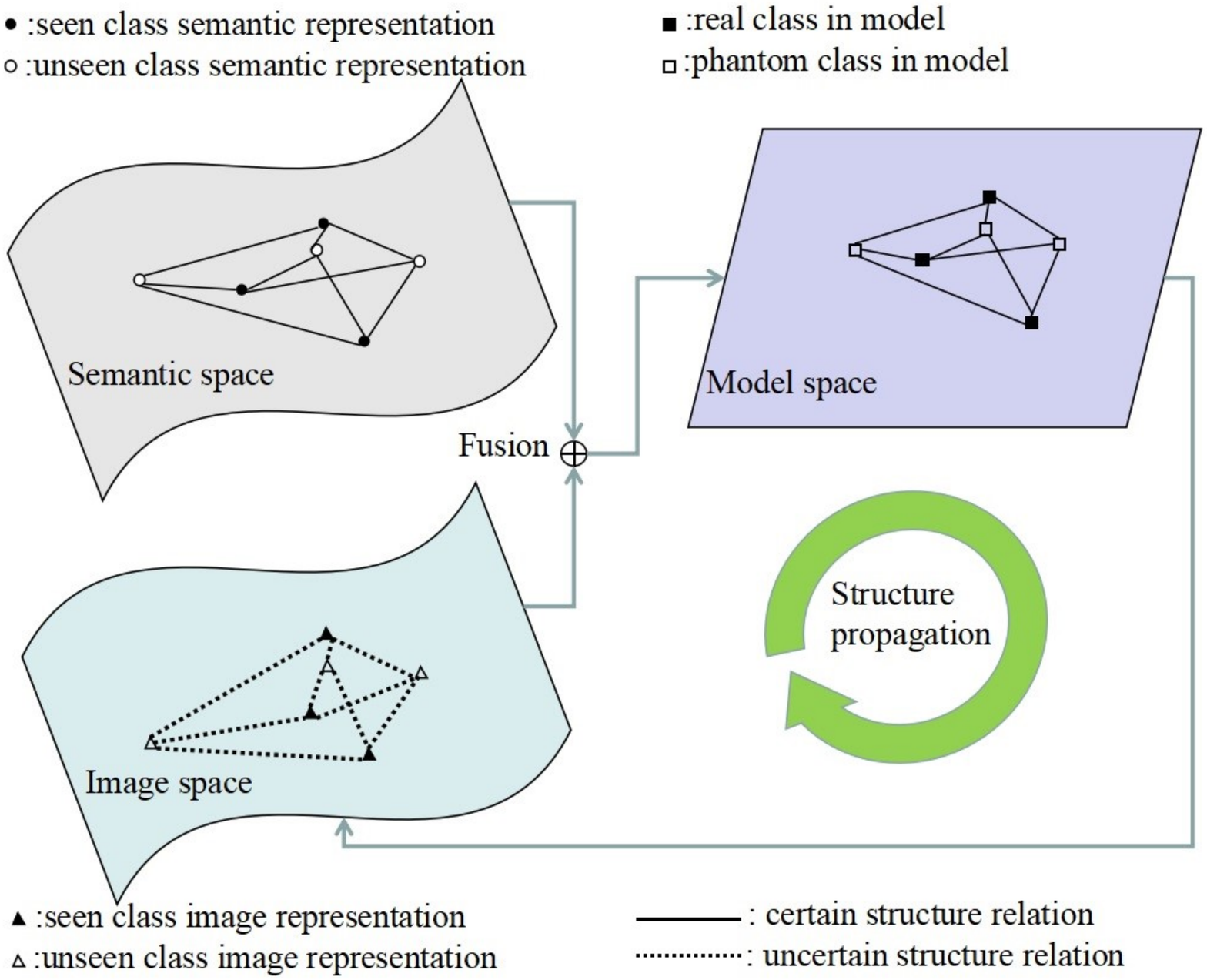}
\end{center}
\vspace{-0.2in}
 \caption{The illustration of structure propagation for zero-shot learning.}
  \label{figmot}
 \end{figure*}

In this paper, our motivation is inspired by structure fusion \cite{Lin2017Dynamic} \cite{Lin2017275} \cite{Lin20161} \cite{Lin2014146} \cite{7268821} \cite{7301305} \cite{Lin20131286}for jointly dealing with two challenges. The intrinsic manifold structure is crucial for object classification. However, in fact, we only can attain the observation data of the manifold structure, which can represent different aspects of the intrinsic manifold structure. For recovering or approximating the intrinsic structure, we can fuse various manifold structures from observation data. Based on the above idea, we try to capture different manifold structures in image and semantic space for improving the recognition performance of unseen classes in ZSL. We view the weighted graph of object classes in semantic or image space as the different manifold structure. In the weighted graph of semantic space (the manifold structure of semantic classes), nodes are corresponding to semantic representations (e.g. attributes\cite{farhadi2009describing}, word vectors \cite{Mikolov2013Distributed}, GloVe \cite{Pennington2014Glove} or Hierarchical embeddings \cite{7298911}) of object classes and these weights of edges describe the distance or similarity relationship of nodes. In the weighted graph of image space(the manifold structure of image classes), it is difficulty to obtain some certain nodes and weights because we do not know labels of unseen classes. Therefore, we expect to construct the compatibility function for predicting labels of unseen classes by building the manifold structure of image classes. On the other end, we attempt to find the relevance between the manifold structure of semantic classes and that of image classes in model space for encoding the influence between the negative and positive transfer, and further make the better compatibility function for classifying unseen class objects. Finally, we iterate the above process to converge the stable state for obtaining the discriminative performance, and in this iteration process manifold structure can be propagated on unseen classes. Model space corresponding to visual appearances is the jointed projection space of semantic space and image space, and can preserve the respective manifold structure. Therefore, we respectively define phantom object classes (the coordinates of classes in the model space are optimized to achieve the best performance of the resulting model for the real object classes in discriminative tasks \cite{Changpinyo2016}.) and real object classes corresponding to all classes in model space. Figure \ref{figmot} illustrates the idea of the proposed method conceptually.

In our main contribution, a novel idea is to recover or approximate the intrinsic manifold structure from seen classes to unseen classes by fusing the different space manifold structure for handling the challenging unseen classes recognition. Specifically, we demonstrate how to construct the projected manifold structure for real and phantom class in model space, and how to constrain the compatibility function and the relationship of the manifold structure for the positive structure propagation. In the experiment, we evaluate SP on four benchmark datasets for ZSL. Experimental results are promising for improving the recognition performance of unseen class objects.

\section{Related Works}

ZSL can bridge the gap among the different domains to recognize unseen class objects by semantic information, which includes the class label and usually can be called the semantic embedding of class labels. These semantic embeddings can come from vision (attributes \cite{lampert2009learning}) and language information (text \cite{Socher2013Zero}) by the manual annotation \cite{Akata2013Label}, machine learning \cite{Yu2013Designing}or data mining \cite{Elhoseiny2013Write}. In term of the transformation relationship of different embedding, recent ZSL methods mainly fall into linear embedding, nonlinear embedding and similarity embedding.

Linear embedding implements the linear transformation method among different embedding spaces for learning the relevance between unseen class objects and class labels. In classical methods, the first step maps  image feature to semantic space, and then the second step recognizes image object by class labels in the semantic space \cite{Akata2013Label} \cite{7298911}. In recent methods,the above steps are combined into a unified framework. Especially, some representation methods, which are max-margin learning learning label embedding \cite{Li2016Semi}, ranking objective\cite{Frome2013DeViSE}, weighted approximate ranking objective \cite{Akata2016Label}, full weight to the top of the ranked list \cite{7298911}and risk minimization formulation with regularization term \cite{romera2015embarrassingly}, can obtain the compatibility of latent space by addressing image and class bi-linear embedding for attaining the better recognition performance of unseen class objects.

Nonlinear embedding can realize the nonlinear mapping of the embedding space for building the compatibility function or classifier. There mainly are three ways for constructing the nonlinear mapping. The first way is a piece-wise linear compatibility for modeling the different characteristic of the embedding \cite{7780384}, and is convenience for computing the transformation matrix of the nonlinear compatibility function in the cross-domain. The second way is a nonlinear hyperbolic tangent activation for learning from image to semantic space of words \cite{Socher2013Zero} or computing inner product of hypothetical space \cite{7505654}, and is suitable for the threshold transformation in the cross-domain. The third way is a kernel function between two images for defining the discriminant function of the intra-modal label transfer \cite{7505654}, and is fit for space metric in the same domain.

\begin{figure*}[ht]
  \begin{center}
\includegraphics[width=0.9\linewidth]{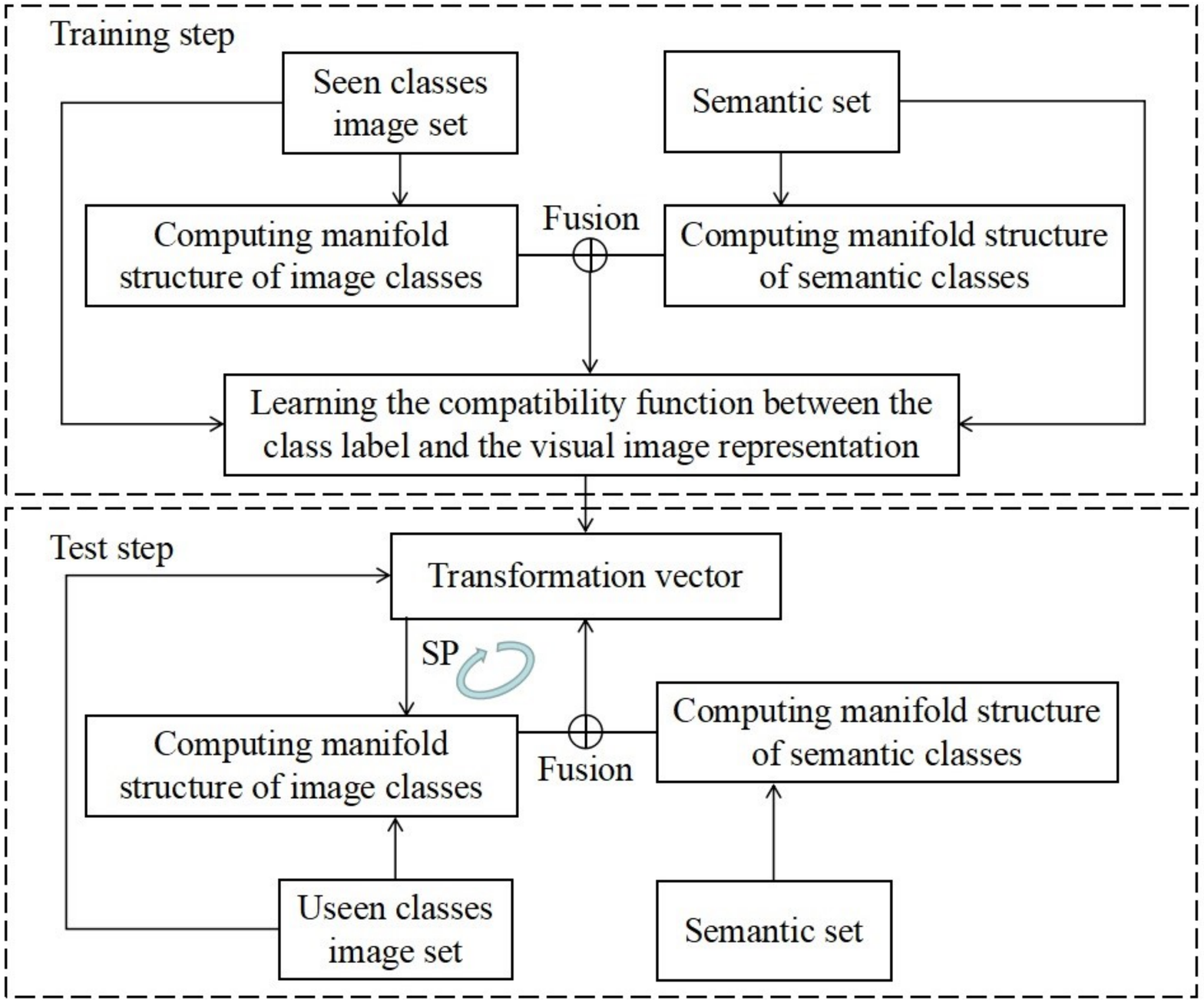}
\end{center}
\vspace{-0.2in}
 \caption{The flowchart of structure propagation (SP) for zero-shot learning (ZSL).}
  \label{figmot1}
 \end{figure*}

Different from the above embeddings, similarity embedding builds the classifier by the similarity metrics, which mostly include structure learning or class-wise similarities. By structure learning, similarity embedding learns a joint latent space in cross domain for fitting each sample by dictionary learning and improving the recognition performance by bilinear classifier \cite{7781018} \cite{zhang2015zero}. Via class-wise similarities, unseen classes associate with seen classes for enhancing the unseen object classification and the domain compatibility \cite{Fu2015Zero} \cite{Mensink2014COSTA} \cite{Changpinyo2016}. Recently, dual visual-semantic mapping paths\cite{Li2017Paths} can capture and refine the semantic space manifold structure (it can be described by the similarity metric) to enhance the transfer ability of visual-semantic mapping for unseen classes classification. In our approach, the similarity metric is extended from semantic space to image space, we attempt to find the relationship of similarities (manifold structure in the different space) for constraining the compatibility function, and further capture to the positive structure propagation for the significantly improvement of the unseen object classification. Figure \ref{figmot1} shows a flowchart for describing the different steps of the proposed method.

\section{Structure propagation}
In ZSL, we have training data set $\mathscr{D}=\{(x_{n}\in R^{D},y_{n})\}_{n=1}^{N}$, in which $x_{n}$ is image representation (it can be extracted based on deep model, and the detail is described in section 4.2.) and $y_{n}(n=1,...,N)$ is the class label in the seen class set $\mathscr{S}=\{s|s=1,...,S\}$. We can denote the unseen class set as $\mathscr{U}=\{u|u=S+1,...,S+U\}$. $a_{c}\in R_{D}$ is the linear transformation vector of the $c \in \{\mathscr{S}\bigcup\mathscr{U}\}$ class. Table \ref{table1} lists the important notations used in this paper.

\begin{table}[!ht]
\small
\begin{center}
\renewcommand{\arraystretch}{1.3}
\caption{List of mathematical notations.}
\label{table1}
\newcommand{\tabincell}[2]{\begin{tabular}{@{}#1@{}}#2\end{tabular}}
\begin{tabular}{cp{8cm}p{8cm}}
\hline
\bfseries Notation & \bfseries Description \\
\hline
$\mathscr{D}=\{(x_{n}\in R^{D},y_{n})\}_{n=1}^{N}$  & Training data set includes the image representation $x_{n}$ and the class label $y_{n}$ in $n$th sample \\
$\mathscr{S}=\{s|s=1,...,S\}$   & The class label space of seen classes\\
$\mathscr{U}=\{u|u=S+1,...,S+U\}$ & The class label space of unseen classes \\
$c \in \{\mathscr{S}\bigcup\mathscr{U}\}$ & The label of any class in $\mathscr{S}\bigcup\mathscr{U}$\\
$a_{c}$  & The transformation vector of the linear model or any real class $c$ representation in model space\\
$v_{s}$  & The phantom class representation corresponding to the seen class $s$ in model space\\
$v_{u}$  & The phantom class representation corresponding to the unseen class $u$ in model space\\
$b_{s}$  & The semantic representation of the seen class $s$ \\
$b_{u}$  & The semantic representation of the unseen class $u$ \\
$x_{s}$  & The image representation of the seen class $s$ \\
$x_{u}$  & The image representation of the unseen class $u$ \\
$w_{su}^{(b)}$ & Similarity between the seen class $s$ and the unseen class $u$ in semantic representation $b$ space\\
$w_{su}^{(x)}$ & Similarity between the seen class $s$ and the unseen class $u$ in semantic representation $x$ space\\
\hline
\end{tabular}
\end{center}
\end{table}

\subsection{Classification model and manifold structure}
We construct a pair-wise linear classifier \cite{Changpinyo2016}in the visual image feature space, and determinate a estimated label $\hat{y}$ to a feature $x$ by the following formula.
 \begin{align}
\label{linear_model}
\begin{aligned}
\hat{y}=\arg\max_{c}a_{c}^{T}x,
 \end{aligned}
\end{align}
here, $a_{c}\in R^{D}$ is not only the transformation vector of the feature $x$, but also the representation of the class $c$ in model. In other words, the above formula can describe the pair-wise linear relation between the feature space and the class label space for characterizing the class representation in the model.

To measure the manifold structure, we can compute the similarity of the related representation in the homogeneous space, which has the same scale and metric. To this end we respectively build a bipartite graph between unseen classes and seen classes in semantic space and image space (this space includes all image representations). In these bipartite graphes, nodes are corresponding to unseen classes or seen classes, and weights of these nodes connect unseen classes with seen classes. Because we focus on the transfer relation between unseen classes and seen classes, no connection exists in unseen classes or seen classes. Supposing $G_{b}<V_{b},E_{b}>$ can denote the manifold structure of semantic classes. Here, $V_{b}=V_{bs}\bigcup V_{bu}$ and $\emptyset=V_{bs}\bigcap V_{bu}$. $E_{b}$ includes connections between $V_{bs}$(seen classes set in semantic space) and $V_{bu}$(unseen classes set in semantic space). Therefore, similarity is regarded as the weight between nodes, which can be defined as following.
 \begin{align}
\label{weight1}
\begin{aligned}
w_{su}^{(b)}=\frac{\exp (-d(b_{s},b_{u}))}{\sum_{u=1}^{U}\exp (-d(b_{s},b_{u}))},
 \end{aligned}
\end{align}
here, $b_{s}$ is the semantic representation (is the vector feature from different semantic sources, and the detail is described in section 4.2.) of the seen class $s$, and $b_{u}$ is the semantic representation of the unseen class $u$. $w_{su}^{(b)}$ is the weight (the similarity ) between the seen class $s$ and the unseen class $u$ in semantic representation $b$ space. $d(b_{s},b_{u})$ is a distance metric \cite{Changpinyo2016}, and can be defined as following.
 \begin{align}
\label{distance1}
\begin{aligned}
d(b_{s},b_{u})=(b_{s}-b_{u})^{T}\Sigma_{b}^{-1}(b_{s}-b_{u}),
 \end{aligned}
\end{align}
here, $\Sigma_{b}=\sigma_{b}I$ can be learned from the semantic representation by cross-validation (We alternately divide the training classes set into two part in according with the proportion between the training classes set and the test classes set. One part is to learn the model, and anther is to validate the model. We give the range of $\sigma_{b}$, which is form $2^{-5}$ to $2^{5}$, and select the parameter corresponding to the best result as the value of $\sigma_{b}$.)

Like the manifold structure of semantic classes, we build a bipartite graphes  $G_{x}<V_{x},E_{x}>$ for the manifold structure of image classes. Here, $V_{x}=V_{xs}\bigcup V_{xu}$ and $\emptyset=V_{xs}\bigcap V_{xu}$. $E_{x}$ includes the connections between $V_{xs}$(seen classes set in image space) and $V_{xu}$(unseen classes set in image space). In term of (\ref{weight1}) and (\ref{distance1}), we can define the weight on $G_{x}$ as following.
 \begin{align}
\label{weight2}
\begin{aligned}
w_{su}^{(x)}=\frac{\exp (-d(x_{s},x_{u}))}{\sum_{u=1}^{U}\exp (-d(x_{s},x_{u}))},
 \end{aligned}
\end{align}
 \begin{align}
\label{distance2}
\begin{aligned}
d(x_{s},x_{u})=(x_{s}-x_{u})^{T}\Sigma_{x}^{-1}(x_{s}-x_{u}),
 \end{aligned}
\end{align}
here, $\Sigma_{x}=\sigma_{x}I$ can be learned from the image representation by cross-validation (It is the same procedure like $\sigma_{b}$ learning.). In image space, the differentiation compared with the semantic space is that $x_{u}$ is not determined because of unseen classes, while $x_{s}$ can be obtained from training data by computing the mean value of the seen class. The way to produce the center of the class as a representation is simple for convenient computation, and it is reasonable to preserve the base characteristic of image representation according with the distribution of the same class. $x_{u}$ can be attained by pre-classification of unseen classes (the detail in the next section).

In (\ref{linear_model}), $a_{c}$ is the transformation vector, and also is the class representation in model space. In (\ref{weight1}), $b_{s}$ and $b_{u}$ is the class representation in semantic space. In (\ref{weight2}), $x_{s}$ and $x_{u}$ is the class representation in image space. We expect to construct the link among these space by $v_{s}$ and $v_{u}$, which are respectively the phantom class of seen or unseen classes in model. For preserving the manifold structure of two bipartite graphes and aligning the image, the semantic and the model space, we build the optimization formula under the condition of the distortion error minimization, which is defined as following.
 \begin{align}
\label{distortion}
\begin{aligned}
(a_{c},v_{u},\vec{\beta})=\arg \min_{a_{c},v_{u},\vec{\beta}} &\|a_{c}-\sum_{u=1}^{U}\vec{\beta}^{T}\left[
\begin{matrix}
w_{su}^{(x)}&w_{su}^{(b)}\\
\end{matrix}
\right]^{T}v_{u}\\
&-\sum_{s=1}^{S}\vec{\gamma}^{T}\left[
\begin{matrix}
w_{ss}^{(x)}&w_{ss}^{(b)}\\
\end{matrix}
\right]^{T}v_{s}\|_{2}^{2},\\
s.t.~~~~&\vec{\beta}^{T}\vec{\mathbf{1}}=1,\vec{\gamma}^{T}\vec{\mathbf{1}}=1,0\leq\beta_{i}\leq1,0\leq\gamma_{i}\leq1~~~~(i=1,2)
 \end{aligned}
\end{align}
here,$\vec{\beta}=\left[
\begin{matrix}
\beta_{1} &\beta_{2}\\
\end{matrix}
\right]^{T}$, $\vec{\gamma}=\left[
\begin{matrix}
\gamma_{1} &\gamma_{2}\\
\end{matrix}
\right]^{T}$,and$\vec{\mathbf{1}}=\left[
\begin{matrix}
1 &1\\
\end{matrix}
\right]^{T}$. Because no connection exists between unseen classes or seen classes in tow bipartite graphes, $w_{ss}^{(b)}=0$ and $w_{ss}^{(x)}=0$. Therefore, (\ref{distortion}) can be transformed into the following formula.
\begin{align}
\label{distortion1}
\begin{aligned}
(a_{c},v_{u},\vec{\beta})=&\arg \min_{a_{c},v_{u},\vec{\beta}} \|a_{c}-\sum_{u=1}^{U}\vec{\beta}^{T}\left[
\begin{matrix}
w_{su}^{(x)}&w_{su}^{(b)}\\
\end{matrix}
\right]^{T}v_{u}\|_{2}^{2},\\
s.t.~~~~~~& \vec{\beta}^{T}\vec{\mathbf{1}}=1,0\leq\beta_{i}\leq1~~~~(i=1,2)
 \end{aligned}
\end{align}
The analytical solution of (\ref{distortion1}) can find the relation between $a_{c}$ and $v_{u}$.
\begin{align}
\label{distortion2}
\begin{aligned}
a_{c}=&\sum_{u=1}^{U}\vec{\beta}^{T}\left[
\begin{matrix}
w_{su}^{(x)}&w_{su}^{(b)}\\
\end{matrix}
\right]^{T}v_{u},\\
s.t.~~~~~~& \vec{\beta}^{T}\vec{\mathbf{1}}=1,0\leq\beta_{i}\leq1~~~~(i=1,2)
 \end{aligned}
\end{align}
here, $\forall c\in \{1,2,...,S+U\}$.

\subsection{Phantom classes and structure relation learning}
For obtaining phantom class $v_{u}(u=1,...,U)$ and the manifold structure of the weight coefficient vector $\beta$, we further reformulate the optimization formula for one-versus-other classifier \cite{Changpinyo2016}.
\begin{align}
\label{opt1}
\begin{aligned}
&(v_{1},...,v_{U},\vec{\beta})=\arg \min_{v_{1},...,v_{U},\vec{\beta}}\sum_{c=1}^{S}\sum_{n=1}^{N}\ell(x_{n},\mathbb{I}_{y_{n},c},a_{c})+\frac{\lambda}{2}\sum_{c=1}^{S}\|a_{c}\|_{2}^{2},\\
&s.t.~~~~~~a_{c}=\sum_{u=1}^{U}\vec{\beta}^{T}\left[
\begin{matrix}
w_{su}^{(x)}&w_{su}^{(b)}\\
\end{matrix}
\right]^{T}v_{u},\vec{\beta}^{T}\vec{\mathbf{1}}=1,0\leq\beta_{i}\leq1~~~~(i=1,2)
 \end{aligned}
\end{align}
here, the first term of formula (\ref{opt1}) is the squared hinge loss, which can be defined as $\ell(x_{n},\mathbb{I}_{y_{n},c}, a_{c})=\max (0,1-\mathbb{I}_{y_{n},c}a_{c}x_{n})$. $\mathbb{I}_{y_{n},c}\in \{-1,1\}$ determinates whether or not $y_{n}=c$. The second term of formula (\ref{opt1}) is $a_{c}$ of a regularization tern, which avoids over-fitting problem on the pair-wise linear classifier for modeling the relationship between the class label and the image representation. However, in formula (\ref{opt1}), we can not determinate the value of $w_{su}^{(x)}$, which can be computed by the certain class label, because the class label of unseen classes can not be got in image space. Therefore, we set the initial value of $w_{su}^{(x)}$ to 0. In other words, we do not consider the manifold structure of the image class in the initial state, and then can complete the optimization of formula (\ref{opt1}) by
solving the quadratic programming problem. When we can categorize the unseen class, and obtain the image class representation by computing the mean value of the same class image representation, in the next iteration computation the updated $w_{su}^{(x)}$  can be considered into the optimization of formula (\ref{opt1}). Although the manifold structure of the image class can be successfully leaded into formula (\ref{opt1}), its influence could have two aspects for recognizing the unseen class. One is the positive effect, which is the mostly correct classification of unseen classes because of the positive structure propagation in each iteration. The other is the negative role, which is the mainly uncorrect classification of unseen classes due to the negative structure propagation. The former situation is our expectation. Therefore, we reformulate formula (\ref{opt1}) by added the third term as following.
\begin{align}
\label{opt2}
\begin{aligned}
(v_{1},...,v_{U},\vec{\beta})=&\arg \min_{v_{1},...,v_{U},\vec{\beta}}\sum_{c=1}^{S}\sum_{n=1}^{N}\ell(x_{n},\mathbb{I}_{y_{n},c},a_{c})+\frac{\lambda}{2}\sum_{c=1}^{S}\|a_{c}\|_{2}^{2}\\
&+\frac{\gamma}{2}\|\beta_{1}W^{x}-\beta_{2}W^{b}\|_{2}^{2},\\
&s.t.~~~~~~a_{c}=\sum_{u=1}^{U}\vec{\beta}^{T}\left[
\begin{matrix}
w_{su}^{(x)}&w_{su}^{(b)}\\
\end{matrix}
\right]^{T}v_{u},\\
&\vec{\beta}^{T}\vec{\mathbf{1}}=1,0\leq\beta_{i}\leq 1~~~~(i=1,2)
 \end{aligned}
\end{align}
here, $w_{su}^{x}$ is the element of the matrix $W^{x}$, and $w_{su}^{b}$ is the element of the matrix $W^{b}$. The third term of formula (\ref{opt2}) is the constraint of the manifold structure similarity for preventing the negative structure propagation in image space. The alternating optimization can be implemented for minimizing the formula (\ref{opt2}) with respect to $\{v_{u}\}_{u=1}^{U} $ and $\vec{\beta}$ by solving the quadratic programming problem.

To depict the whole process of the structure propagation mechanism, we show the pseudo code of the proposed SP algorithm in Algorithm \ref{algSP}, which includes three steps. The first step (line 1) computes the similarity matrix to represent the manifold structure of semantic classes. The second step (line 2) initializes the similarity matrix related to the manifold structure of image classes. The third step includes phantom classes $\{v_{u}\}_{u=1}^{U} $, weight coefficients $\vec{\beta}$  and the classification of unseen object classes updating by each iteration (from line 3 to line 9). Structure propagation can be completed by the whole iteration computation.

\begin{algorithm}[ht]
  \caption{The pseudo code of the SP algorithm}
 \begin{algorithmic}[1]
 \label{algSP}
\renewcommand{\algorithmicrequire}{\textbf{Input:}}
\renewcommand{\algorithmicensure}{\textbf{Output:}}
\renewcommand{\algorithmicreturn}{\textbf{Iteration:}}
   \REQUIRE $\mathscr{D}=\{(x_{n}\in R^{D},y_{n})\}_{n=1}^{N}$,$b_{s}$ and $b_{u}$  (input data)
   \ENSURE $y^{*}_{P}$ ($P$ is the total iteration number )
   \STATE Computes the similarity matrix $W_{(b)}$ on the semantic representation by (\ref{weight1})
   \STATE Setting the similarity matrix $W_{(x)}$ to zero matrix on the image representation
   \FOR {$1<t<P$}
   \STATE Solving $\{v_{u}\}_{u=1}^{U} $ and $\vec{\beta}$ by alternately optimizing (\ref{opt2})
   \STATE Computing $a_{c}$ according to (\ref{distortion2})
   \STATE Computing $\hat{y}$ by (\ref{linear_model}) and obtaining the class label $y^{*}_{t}$ of the unseen class corresponding to the semantic class
   \STATE Computing the mean value of each image class as the image class representation $x_{s}$ and $x_{u}$
   \STATE Computing and updating the similarity matrix $W_{(x)}$ on the image representation by (\ref{weight2})
   \ENDFOR
  \end{algorithmic}
\end{algorithm}

\subsection{Multi-semantic structure fusion}
To address multi-semantic structure fusion, we produce $w_{su}^{b}$ by the linear fusion of multi-semantic structure as following.
\begin{align}
\label{fusion1}
\begin{aligned}
w_{su}^{b}=\vec{\eta}^{T}\left[
\begin{matrix}
w_{su}^{(ba)}&w_{su}^{(bw)}&w_{su}^{(bg)}&w_{su}^{(bh)}\\
\end{matrix}
\right]^{T},
 \end{aligned}
\end{align}
here, $\vec{\eta}=\left[
\begin{matrix}
\beta_{2}&\beta_{3}&\beta_{4}&\beta_{5}\\
\end{matrix}
\right]^{T}$. $w_{su}^{(ba)}$, $w_{su}^{(bw)}$, $w_{su}^{(bg)}$, and $w_{su}^{(bh)}$ are respectively corresponding to attributes (att)\cite{farhadi2009describing}, word vectors(w2v) \cite{Mikolov2013Distributed}, GloVe (glo)\cite{Pennington2014Glove} and Hierarchical embeddings (hie)\cite{7298911}. We can bring formula (\ref{fusion1}) into formula (\ref{opt2}) for handling the multi-semantic structure fusion as following.
\begin{align}
\label{fusion2}
\begin{aligned}
(v_{1},...,v_{U},\vec{\beta})=&\arg \min_{v_{1},...,v_{U},\vec{\beta}}\sum_{c=1}^{S}\sum_{n=1}^{N}\ell(x_{n},\mathbb{I}_{y_{n},c},a_{c})+\frac{\lambda}{2}\sum_{c=1}^{S}\|a_{c}\|_{2}^{2}\\
&+\frac{\gamma}{2}\|\beta_{1}W^{x}-\beta_{2}W^{ba}-\beta_{3}W^{bw}-\beta_{4}W^{bg}-\beta_{5}W^{bh}\|_{2}^{2},\\
s.t.~~~~~~&a_{c}=\sum_{u=1}^{U}\vec{\beta}^{T}\left[
\begin{matrix}
w_{su}^{(x)}&w_{su}^{(ba)}&w_{su}^{(bw)}&w_{su}^{(bg)}&w_{su}^{(bh)}\\
\end{matrix}
\right]^{T}v_{u},\\
&\vec{\beta}^{T}\vec{\mathbf{1}}=1,0\leq\beta_{i}\leq1~~~~(i=1,...,5)
 \end{aligned}
\end{align}
here, $\vec{\beta}=\left[
\begin{matrix}
\beta_{1} &\beta_{2}&\beta_{3}&\beta_{4}&\beta_{5}\\
\end{matrix}
\right]^{T}$,$w_{su}^{(ba)}$, $w_{su}^{(bw)}$, $w_{su}^{(bg)}$, and $w_{su}^{(bh)}$ are respectively the element of $W_{su}^{(ba)}$, $W_{su}^{(bw)}$, $W_{su}^{(bg)}$, and $W_{su}^{(bh)}$. If there are the more semantic information, we can consider these semantic information for ZSL by the similar way of formula (\ref{fusion2}). Like Algorithm \ref{algSP}, (\ref{fusion2}) has the similar optimization solving process for considering multi-semantic information in ZSL.

\subsection{Complexity analysis}
Formula (\ref{opt2}) can be solved by alternately quadratic programming, which of the complexity includes two parts. In the first part, when $\vec{\beta}$ is fixed, formula (\ref{opt2}) is related to $\{v_{u}\}_{u=1}^{U} $ of a quadratic programming problem, which of the complexity is $O(U^{3})$ for the worst \cite{Boyd2013Convex}. In the second part, while $\{v_{u}\}_{u=1}^{U} $ is fixed, formula (\ref{opt2}) is corresponding to $\vec{\beta}$  of a quadratic programming problem, which of the complexity is $O(k^{3})$ ($k$ is the dimension of $\vec{\beta}$)for the worst \cite{Boyd2013Convex}. Given the proposed algorithm SP needs $P$ iterations, it's complexity is $O(PU^{3}+Pk^{3})$.

\section{Experiment}
\subsection{Datasets}
\begin{figure*}[ht]
  \begin{center}
\includegraphics[width=0.8\linewidth]{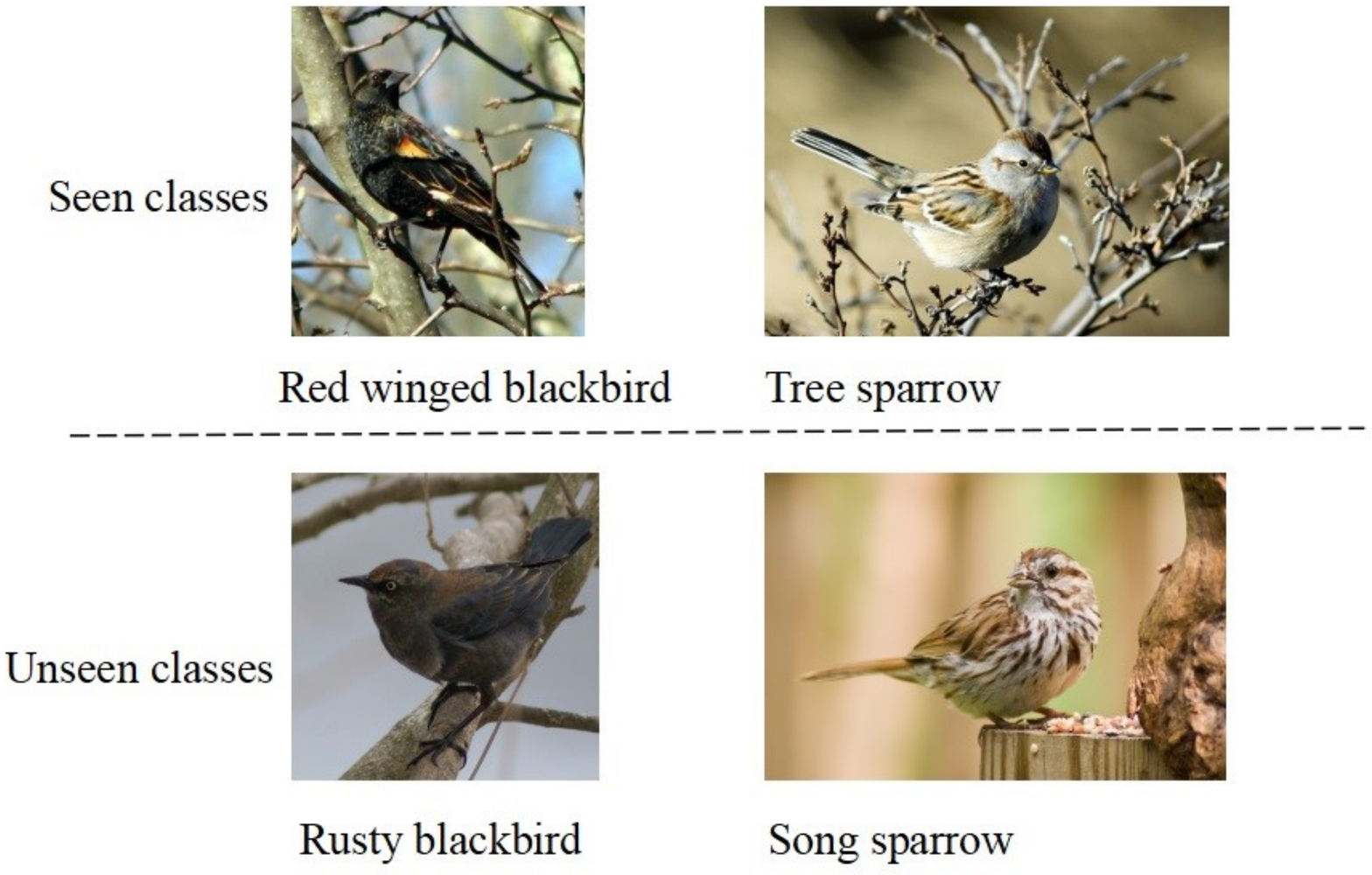}
\end{center}
\vspace{-0.2in}
 \caption{Image examples in CUB-200-2011 Birds database.}
  \label{figmot2}
 \end{figure*}

For evaluating the proposed algorithm SP, we carry out the experiment in four challenging datasets, which are Animals with Attributes (AwA)\cite{lampert2014attribute}, CUB-200-2011 Birds (CUB)\cite{Wah2011The}, Stanford Dogs (Dogs)\cite{Deng2013Fine}, and  SUN Attribute (SUN)\cite{Patterson2014The}. These datasets can be used for fine-grained recognition (CUB and Dogs) or non-fine-grained recognition (AwA and SUN) in ZSL. In semantic space, AwA and CUB respectively are described by att\cite{farhadi2009describing},w2v \cite{Mikolov2013Distributed},glo\cite{Pennington2014Glove} and hie\cite{7298911}, while Dogs is represented by w2v \cite{Mikolov2013Distributed},glo\cite{Pennington2014Glove} and hie\cite{7298911}. SUN is only depicted by att\cite{farhadi2009describing}. Figure \ref{figmot2} shows image examples in CUB-200-2011 Birds database.  Tab.\ref{table2} provides the statistics and the extracted features for these datasets.

\begin{table*}[!ht]
\small
\renewcommand{\arraystretch}{1.3}
\caption{Datasets statistics and the extracted feature in experiments.}
\label{table2}
\begin{center}
\newcommand{\tabincell}[2]{\begin{tabular}{@{}#1@{}}#2\end{tabular}}
\begin{tabular}{lp{2cm}p{2cm}p{2cm}p{2cm}p{2cm}p{1cm}}
\hline
\bfseries Datasets & \bfseries \tabincell{l}{Number of \\seen classes} & \bfseries \tabincell{l}{Number of \\unseen classes} & \bfseries \tabincell{l}{Total number \\of images} & \bfseries \tabincell{l}{Semantic feature\\/dimension} &\bfseries \tabincell{l}{Image feature\\/dimension}\\
\hline
AwA  & $40$ &$10$& $30473$ & \tabincell{l}{att/85,\\w2v/400,\\glo/400,\\hie/about 200.} & \tabincell{l}{Deep feature based\\ on GoogleNet\cite{7298594}\\/1024}\\
CUB  & $150$ &$50$& $11786$ & \tabincell{l}{att/312,\\w2v/400,\\glo/400,\\hie/about 200.} & \tabincell{l}{Deep feature based\\ on GoogleNet\cite{7298594}\\/1024}\\
Dogs  & $85$ &$28$& $19499$ & \tabincell{l}{N/A,\\w2v/400,\\glo/400,\\hie/about 200.} & \tabincell{l}{Deep feature based\\ on GoogleNet\cite{7298594}\\/1024}\\
SUN  & $645$ &$72$& $14340$ & \tabincell{l}{att/102,\\N/A,\\N/A,\\N/A.} & \tabincell{l}{Deep featurebased\\ on GoogleNet\cite{7298594}\\/1024}\\
\hline
\end{tabular}
\end{center}
\end{table*}
\subsection{Image and semantic feature}
In our SP \footnote{source code:https://github.com/lgf78103/Structure-propagation-for-zero-shot-learning.} method, image feature and semantic feature are the main support for modeling ZSL. Deep learning feature can learn the discriminative characteristic of objects based on large scale database. In addition, for conveniently comparing with the state-of-art methods, we adopt image feature provided by \cite{7298911}. In one word, image feature is the outputs (1024 dimension feature vector) of the pre-trained GoogleNet\cite{7298594}, which can process the whole image as inputs. These images are not pre-processed in any way. In the semantic space, there are four ways to extract the related feature. The first way is the distinguishing vector feature of objects (att) from attributes \cite{farhadi2009describing} by human annotation and judgment, which have been completed for collecting data on AwA, CUB, and SUN except Dogs. The second ways is word vectors(w2v) based on a two-layer neural network to predict words through a text document\cite{Mikolov2013Distributed}. The third way is GloVe (glo) based on co-occurrence statistics of words from a large unlabel text corpora \cite{Pennington2014Glove}. The forth way is hierarchical embeddings (hie) based on vectorial class structure from the class hierarchical relationship such as WordNet\cite{7298911}\cite{Miller2002WordNet}. The w2v, glo, and hie are also provided by \cite{7780384} to facilitate contrast to the state-of-art methods. In addition, we can extend the other types of  visual features in the proposed method, as will be studied for feature fusion in future work.

\subsection{Comparison methods}
In this paper, there are three methods as the baseline for comparing with the proposed SP method because of the semantic structure mining. The first method is structured joint embedding (SJE) \cite{7298911}, which can build the bilinear compatibility function with the consideration of the structured output space for predicting the label of the unseen class. The second method is latent embedding model (LatEm)\cite{7780384},which can construct the pair-wise bilinear (nonlinear) compatibility function according to model number selection for recognizing unseen classes. The third method is synthesized classifiers (SynC)\cite{Changpinyo2016}, which can make nonlinear compatibility function with manifold structure in semantic space for combining the base classifier in ZSL.
\subsection{Classification and validation protocols}
Classification accuracy is average value of all test class accuracy in each database. Because the learned model involves four parameters, which are $\lambda$, $\gamma$ , $\sigma_{b}$ and $\sigma_{x}$ (respectively are in formula (\ref{distance1}) and (\ref{distance2})) in formula (\ref{opt2}). We alternately divide the training classes set into two part in according with the proportion between the training classes set and the test classes set. One part is to learn the model, and anther is to validate the model. Firstly, we set $\sigma_{b}$ and $\sigma_{x}$ to 1, and obtain $\gamma$ and $\lambda$ corresponding to the best result in $\gamma$ (form $2^{-24}$ to $2^{-9}$) and $\lambda$ (form $2^{-24}$ to $2^{-9}$) by cross validation. Secondly, we learn $\sigma_{b}$ and $\sigma_{x}$ corresponding to the best result in $\sigma_{b}$ and $\sigma_{x}$ (form $2^{-5}$ to $2^{5}$) by cross validation.
\subsection{AwA}
\begin{figure*}[ht]
  \begin{center}
\includegraphics[width=1\linewidth]{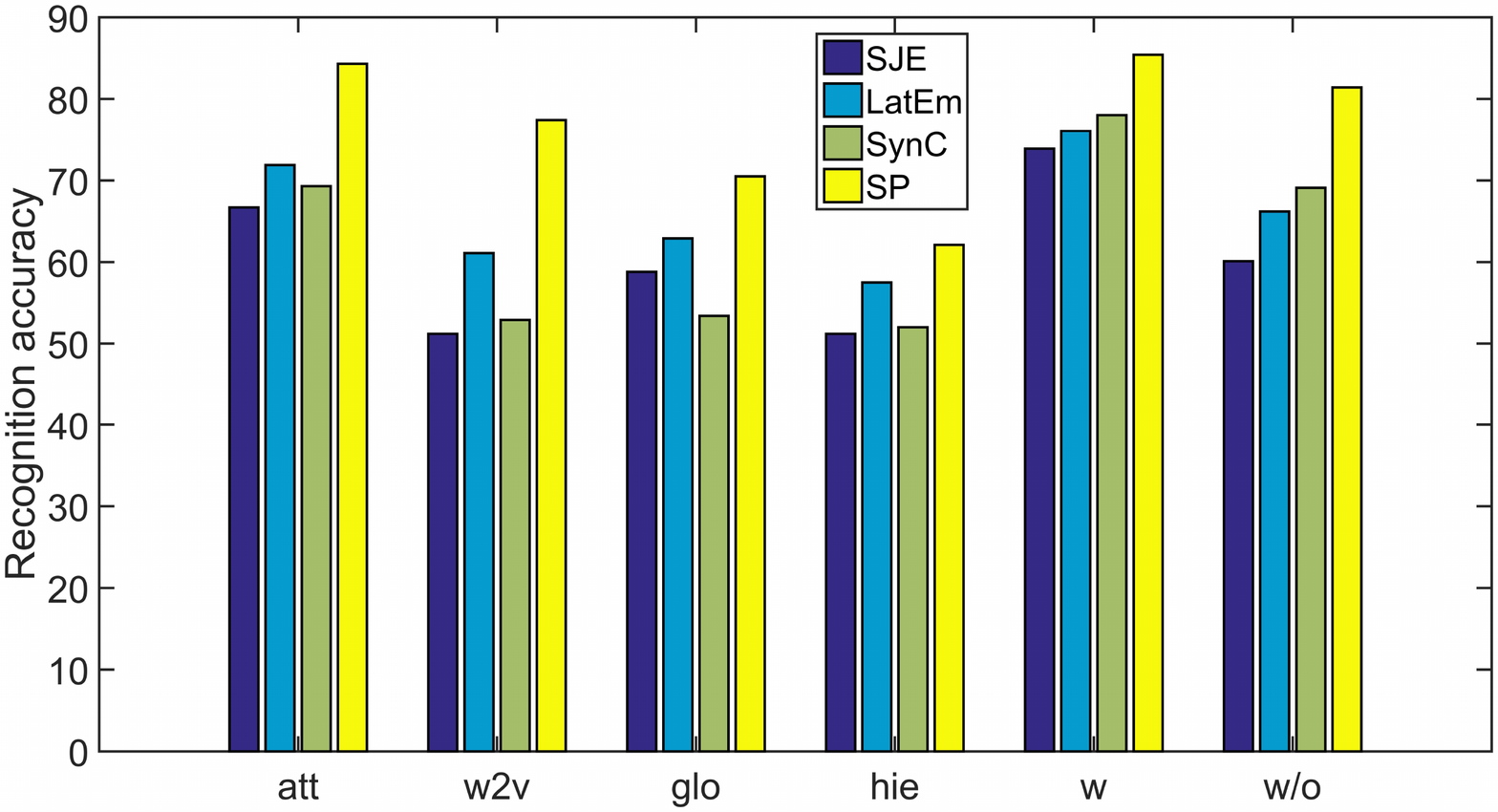}
\end{center}
\vspace{-0.2in}
 \caption{Comparison of SP method with SJE\cite{7298911}, LatEm\cite{7780384} and SynC\cite{Changpinyo2016} in att, w2v, glo, hie, w (the fusion includes att, w2v, glo and hie) and w/o (the fusion contains w2v, glo and hie), average per-class Top-1 accuracy of unseen classes is reported based on the same data configurations, same images and semantic features in AwA.}
  \label{fig2}
 \end{figure*}

Animals with Attributes (AwA)\cite{lampert2014attribute} is popularly used for ZSL. The characteristic of this dataset is the large scale and the small categories. We extract the deep feature of the image based on the pre-trained GoogleNet\cite{7298594}, the continuous vector (att) for clustering attributes in semantic classes \cite{farhadi2009describing}, the word vector (w2v)for describing words in the specific context\cite{Mikolov2013Distributed}, the glover vector (glo) for gathering the co-occurrence words statistics in the given document\cite{Pennington2014Glove}, and the hierarchy vector (hie) for measuring the distance of the hierarchy structure in semantic classes \cite{7298911}. We have two configurations for ZSL in AwA. One is the comparison SP with the state-of-art methods in each semantic space, the other is the comparison experiment of the fusion methods in multi-semantic space. Figure \ref{fig2} indicates the experimental comparison of the different method in the various semantic space of AwA. Table \ref{table3} shows the performance of the structure propagation (the proposed SP method) greatly outperforms that of other three methods. The highest and the lowest improvement of SP are respectively $26.2\%$ and $10.9\%$ to compare with SJE, $16.3\%$ and $4.6\%$ to contrast to LatEm, or $24.5\%$ and $10.1\%$ to contradistinguish SynC. Table \ref{table4} demonstrates the performance of the SP fusion method is better than that of other three fusion methods. Specifically, when the fusion includes the supervised (att) and unsupervised (w2v, glo,and hie) semantic representation, the accuracy of SP can be increased by $11.5\%$ than SJE, $9.3\%$ than LatEm, and $7.4\%$ than SynC. While the fusion only contains the unsupervised semantic description (w2v, glo,and hie), the precision of SP can be enhanced by $21.3\%$ than SJE, $15.2\%$ than LatEm, and $12.3\%$ than SynC.

\begin{table*}[ht]
\small
\renewcommand{\arraystretch}{1.3}
\caption{Comparison of SP method with SJE\cite{7298911}, LatEm\cite{7780384} and SynC\cite{Changpinyo2016} in each semantic space, average per-class Top-1 accuracy (\%)of unseen classes is reported based on the same data configurations, same images and semantic features in AwA.}
\label{table3}
\begin{center}
\newcommand{\tabincell}[2]{\begin{tabular}{@{}#1@{}}#2\end{tabular}}
\begin{tabular}{lp{2cm}p{2cm}p{2cm}p{2cm}}
\hline
\bfseries  Semantic feature &\bfseries SJE &\bfseries LatEm  &\bfseries SynC  &\bfseries SP\\
\hline
att  & $66.7$ &$71.9$& $69.3$ & $\textbf{84.3}$\\
w2v  & $51.2$ &$61.1$& $52.9$ & $\textbf{77.4}$\\
glo  & $58.8$ &$62.9$& $53.4$ & $\textbf{70.5}$\\
hie  & $51.2$ &$57.5$& $52.0$ & $\textbf{62.1}$\\
\hline
\end{tabular}
\end{center}
\end{table*}

\begin{table*}[ht]
\small
\renewcommand{\arraystretch}{1.3}
\caption{Comparison of SP method with SJE\cite{7298911}, LatEm\cite{7780384} and SynC\cite{Changpinyo2016} for multi-semantic fusion, average per-class Top-1 accuracy (\%)of unseen classes is reported based on the same data configurations, same images and semantic features in AwA. w: the fusion includes att, w2v, glo and hie, while w/o: the fusion contains w2v, glo and hie.}
\label{table4}
\begin{center}
\newcommand{\tabincell}[2]{\begin{tabular}{@{}#1@{}}#2\end{tabular}}
\begin{tabular}{lp{2cm}p{2cm}p{2cm}p{2cm}}
\hline
\bfseries Fusion &\bfseries SJE &\bfseries LatEm  &\bfseries SynC  &\bfseries SP\\
\hline
w  & $73.9$ &$76.1$& $78.0$ & $\textbf{85.4}$\\
w/o  & $60.1$ &$66.2$& $69.1$ & $\textbf{81.4}$\\
\hline
\end{tabular}
\end{center}
\end{table*}

\subsection{CUB}
\begin{figure*}[ht]
  \begin{center}
\includegraphics[width=1\linewidth]{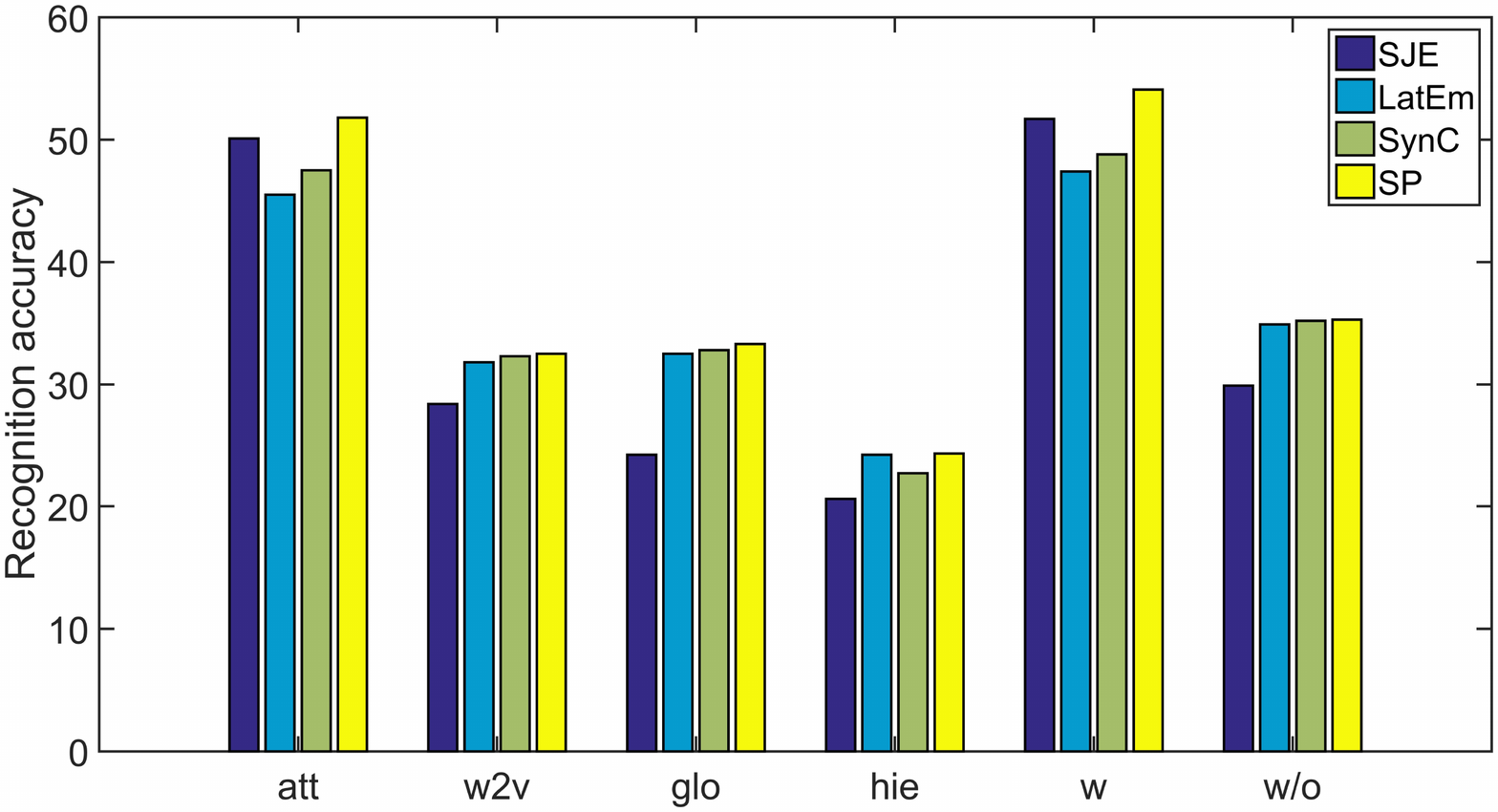}
\end{center}
\vspace{-0.2in}
 \caption{Comparison of SP method with SJE\cite{7298911}, LatEm\cite{7780384} and SynC\cite{Changpinyo2016} in att, w2v, glo, hie, w (the fusion includes att, w2v, glo and hie) and w/o (the fusion contains w2v, glo and hie), average per-class Top-1 accuracy of unseen classes is reported based on the same data configurations, same images and semantic features in CUB.}
  \label{fig3}
 \end{figure*}

CUB-200-2011 Birds (CUB)\cite{Wah2011The} is generally utilized for the fine-grained recognition. The scale of this dataset is smaller than AwA, but there are the more categories in CUB. Like AwA, we extracted the deep feature of images in CUB, and we get att, w2v, glo and hie by the semantic description of CUB. Two configurations are respectively the non-fusion and fusion methods comparison in the single and multi-semantic space. Figure \ref{fig3} indicates the experimental comparison of the different method in the various semantic space of CUB. Table \ref{table5} shows that the performance of the structure propagation (the proposed SP method) outperforms that of other three methods. The highest and the lowest improvement of SP are respectively $9.1\%$ and $1.7\%$ to compare with SJE, $6.3\%$ and $0.1\%$ to contrast to LatEm, or $4.3\%$ and $0.2\%$ to contradistinguish SynC. Table \ref{table6} demonstrates that the performance of the SP fusion method is slightly better than that of other three fusion methods. Specifically, when the fusion includes the supervised (att) and unsupervised (w2v, glo,and hie) semantic representation, the accuracy of SP can be increased by $2.4\%$ than SJE, $6.7\%$ than LatEm, and $5.3\%$ than SynC. While the fusion only contains the unsupervised semantic description (w2v, glo,and hie), the precision of SP can be enhanced by $5.4\%$ than SJE, $0.4\%$ than LatEm, and $0.1\%$ than SynC.

\begin{table*}[ht]
\small
\renewcommand{\arraystretch}{1.3}
\caption{Comparison of SP method with SJE\cite{7298911}, LatEm\cite{7780384} and SynC\cite{Changpinyo2016} in each semantic space, average per-class Top-1 accuracy(\%) of unseen classes is reported based on the same data configurations, same images and semantic features in CUB.}
\label{table5}
\begin{center}
\newcommand{\tabincell}[2]{\begin{tabular}{@{}#1@{}}#2\end{tabular}}
\begin{tabular}{lp{2cm}p{2cm}p{2cm}p{2cm}}
\hline
\bfseries  Semantic feature &\bfseries SJE &\bfseries LatEm  &\bfseries SynC &\bfseries SP\\
\hline
att  & $50.1$ &$45.5$& $47.5$ & $\textbf{51.8}$\\
w2v  & $28.4$ &$31.8$& $32.3$ & $\textbf{32.5}$\\
glo  & $24.2$ &$32.5$& $32.8$ & $\textbf{33.3}$\\
hie  & $20.6$ &$24.2$& $22.7$ & $\textbf{24.3}$\\
\hline
\end{tabular}
\end{center}
\end{table*}

\begin{table*}[ht]
\small
\renewcommand{\arraystretch}{1.3}
\caption{Comparison of SP method with SJE\cite{7298911}, LatEm\cite{7780384} and SynC\cite{Changpinyo2016} for multi-semantic fusion, average per-class Top-1 accuracy(\%) of unseen classes is reported based on the same data configurations, same images and semantic features in CUB. w: the fusion includes att, w2v, glo and hie, while w/o: the fusion contains w2v, glo and hie.}
\label{table6}
\begin{center}
\newcommand{\tabincell}[2]{\begin{tabular}{@{}#1@{}}#2\end{tabular}}
\begin{tabular}{lp{2cm}p{2cm}p{2cm}p{2cm}}
\hline
\bfseries Fusion &\bfseries SJE &\bfseries LatEm  &\bfseries SynC  &\bfseries SP\\
\hline
w  & $51.7$ &$47.4$& $48.8$ & $\textbf{54.1}$\\
w/o  & $29.9$ &$34.9$& $35.2$ & $\textbf{35.3}$\\
\hline
\end{tabular}
\end{center}
\end{table*}

\subsection{Dogs}
\begin{figure*}[ht]
  \begin{center}
\includegraphics[width=1\linewidth]{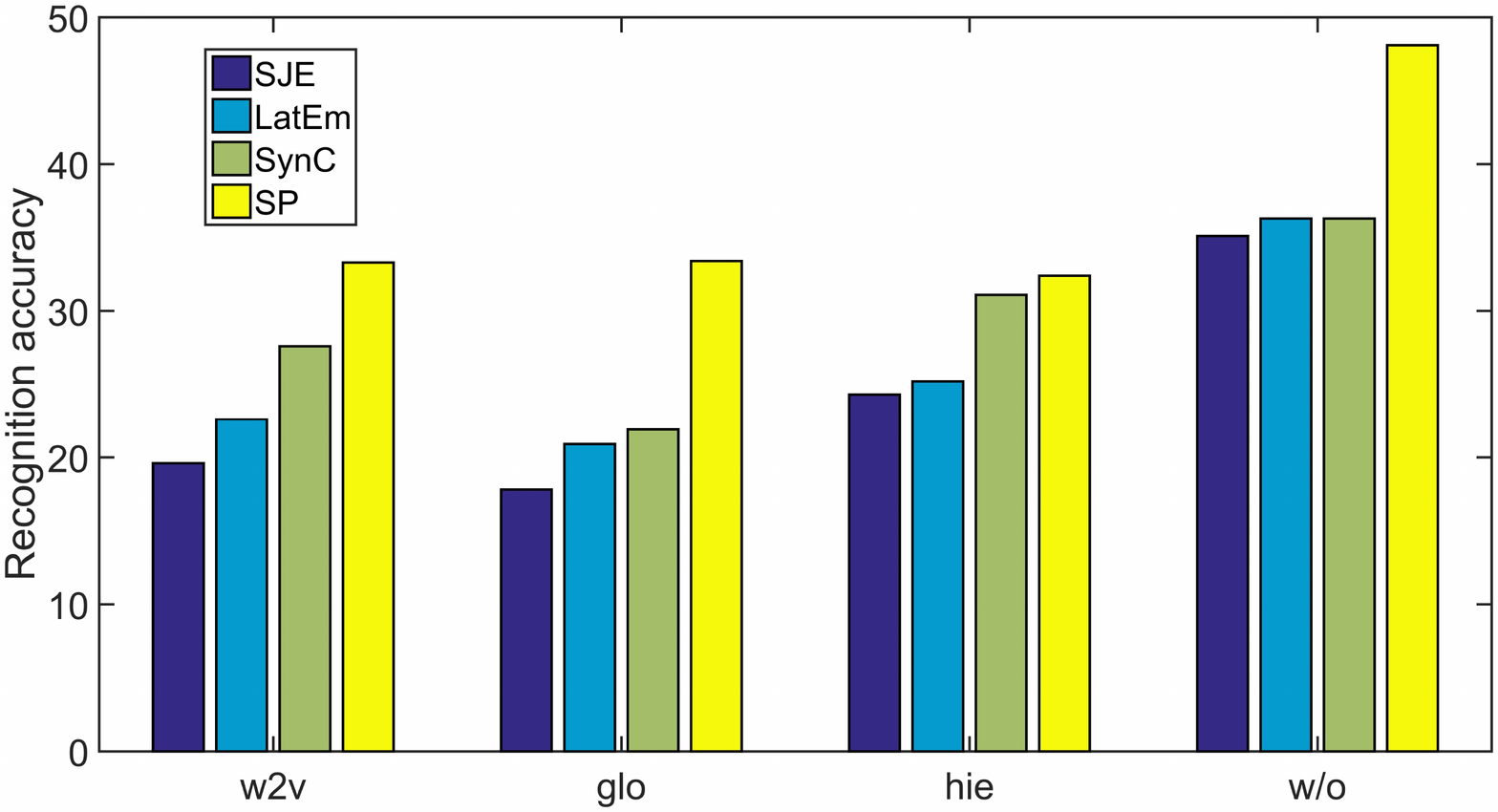}
\end{center}
\vspace{-0.2in}
 \caption{Comparison of SP method with SJE\cite{7298911}, LatEm\cite{7780384} and SynC\cite{Changpinyo2016} in w2v, glo, hie  and w/o (the fusion contains w2v, glo and hie), average per-class Top-1 accuracy of unseen classes is reported based on the same data configurations, same images and semantic features in Dogs.}
  \label{fig4}
 \end{figure*}

Stanford Dogs (Dogs)\cite{Deng2013Fine} is also usually a benchmark dataset for fine-grained recognition. Dogs is middle between AwA and CUB about the scale and the category number. We use the same method for extracting the deep feature of images and semantic class features (w2v, glo and hie). We also carry out the experiment in non-fusion methods in the single semantic space and fusion methods in the multi-semantic space. Figure \ref{fig4} indicates the experimental comparison of the different method in the various semantic space of Dogs.  Table \ref{table7} shows the performance of the structure propagation (the proposed SP method) outperforms that of other three methods. The highest and the lowest improvement of SP are respectively $15.6\%$ and $8.1\%$ to compare with SJE, $12.5\%$ and $7.2\%$ to contrast to LatEm, or $11.5\%$ and $1.3\%$ to contradistinguish SynC. Table \ref{table8} demonstrates the performance of the SP fusion method is obviously better than that of other three fusion methods. Specifically, While the fusion only contains the unsupervised semantic description (w2v, glo,and hie), the precision of SP can be enhanced by $13\%$ than SJE, $11.8\%$ than LatEm, and $11.8\%$ than SynC.

\begin{table*}[ht]
\small
\renewcommand{\arraystretch}{1.3}
\caption{Comparison of SP method with SJE\cite{7298911}, LatEm\cite{7780384} and SynC\cite{Changpinyo2016} in each semantic space, average per-class Top-1 accuracy(\%) of unseen classes is reported based on the same data configurations, same images and semantic features in Dogs.}
\label{table7}
\begin{center}
\newcommand{\tabincell}[2]{\begin{tabular}{@{}#1@{}}#2\end{tabular}}
\begin{tabular}{lp{2cm}p{2cm}p{2cm}p{2cm}}
\hline
\bfseries  Semantic feature &\bfseries SJE &\bfseries LatEm  &\bfseries SynC  &\bfseries SP\\
\hline
att  & $N/A$ &$N/A$& $N/A$ & $N/A$\\
w2v  & $19.6$ &$22.6$& $27.6$ & $\textbf{33.3}$\\
glo  & $17.8$ &$20.9$& $21.9$ & $\textbf{33.4}$\\
hie  & $24.3$ &$25.2$& $31.1$ & $\textbf{32.4}$\\
\hline
\end{tabular}
\end{center}
\end{table*}

\begin{table*}[ht]
\small
\renewcommand{\arraystretch}{1.3}
\caption{Comparison of SP method with SJE\cite{7298911}, LatEm\cite{7780384} and SynC\cite{Changpinyo2016} for multi-semantic fusion, average per-class Top-1 accuracy(\%) of unseen classes is reported based on the same data configurations, same images and semantic features in Dogs. w: the fusion includes att, w2v, glo and hie, while w/o: the fusion contains w2v, glo and hie.}
\label{table8}
\begin{center}
\newcommand{\tabincell}[2]{\begin{tabular}{@{}#1@{}}#2\end{tabular}}
\begin{tabular}{lp{2cm}p{2cm}p{2cm}p{2cm}}
\hline
\bfseries Fusion &\bfseries SJE &\bfseries LatEm  &\bfseries SynC  &\bfseries SP\\
\hline
w  & $N/A$ &$N/A$& $N/A$ & $N/A$\\
w/o  & $35.1$ &$36.3$& $36.3$ & $\textbf{48.1}$\\
\hline
\end{tabular}
\end{center}
\end{table*}

\subsection{SUN}
\begin{figure*}[ht]
  \begin{center}
\includegraphics[width=1\linewidth]{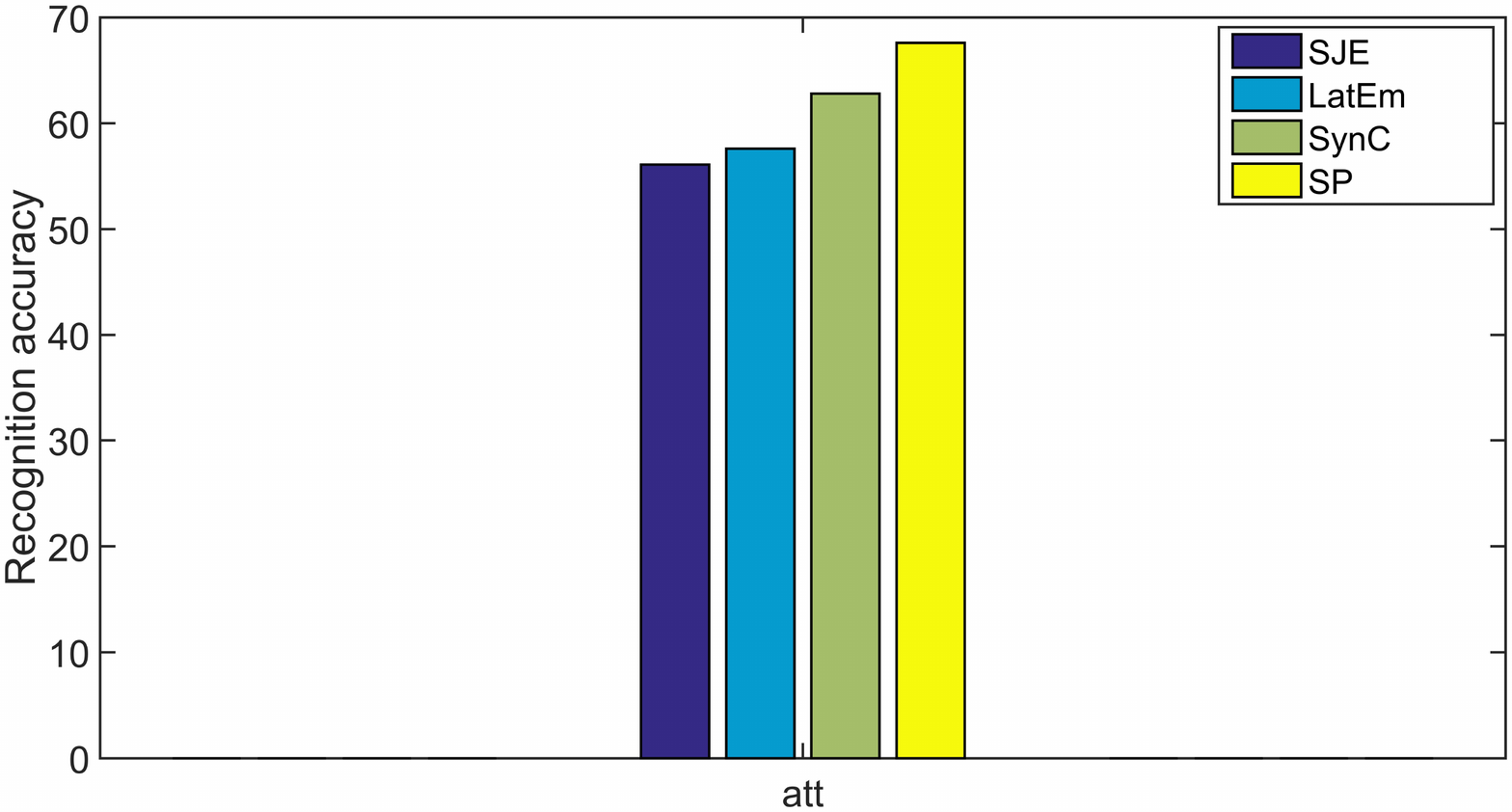}
\end{center}
\vspace{-0.2in}
 \caption{Comparison of SP method with SJE\cite{7298911}, LatEm\cite{7780384} and SynC\cite{Changpinyo2016} in att, average per-class Top-1 accuracy of unseen classes is reported based on the same data configurations, same images and semantic features in SUN.}
  \label{fig5}
 \end{figure*}

SUN Attribute (SUN)\cite{Patterson2014The} is the first large-scale scene attribute dataset. Because scene is greatly more complex than the specific object (e.g. bear, dog, or bird), so it is difficult to find the unsupervised source (e.g. wikipedia for w2v and glo) for precisely describing the scene semantics. Therefore, we only use att for implementing the experiment. Figure \ref{fig5} indicates the experimental comparison of the different method in the attribute semantic space of SUN. Table \ref{table9} demonstrates the performance of the SP method is superior to that of other three methods. Specifically, the accuracy of SP can be enhanced by $11.5\%$ than SJE, $10\%$ than LatEm, and $4.8\%$ than SynC.

\begin{table*}[ht]
\small
\renewcommand{\arraystretch}{1.3}
\caption{Comparison of SP method with SJE\cite{7298911}, LatEm\cite{7780384} and SynC\cite{Changpinyo2016} in attribute semantic space, average per-class Top-1 accuracy(\%) of unseen classes is reported based on the same data configurations, same images and semantic features in SUN.}
\label{table9}
\begin{center}
\newcommand{\tabincell}[2]{\begin{tabular}{@{}#1@{}}#2\end{tabular}}
\begin{tabular}{lp{2cm}p{2cm}p{2cm}p{2cm}}
\hline
\bfseries  Semantic feature &\bfseries SJE &\bfseries LatEm  &\bfseries SynC  &\bfseries SP\\
\hline
att  & $56.1$ &$57.6$& $62.8$ & $\textbf{67.6}$\\
\hline
\end{tabular}
\end{center}
\end{table*}

\subsection{Structure propagation with the iteration}
\begin{figure*}[ht]
  \begin{center}
\includegraphics[width=1\linewidth]{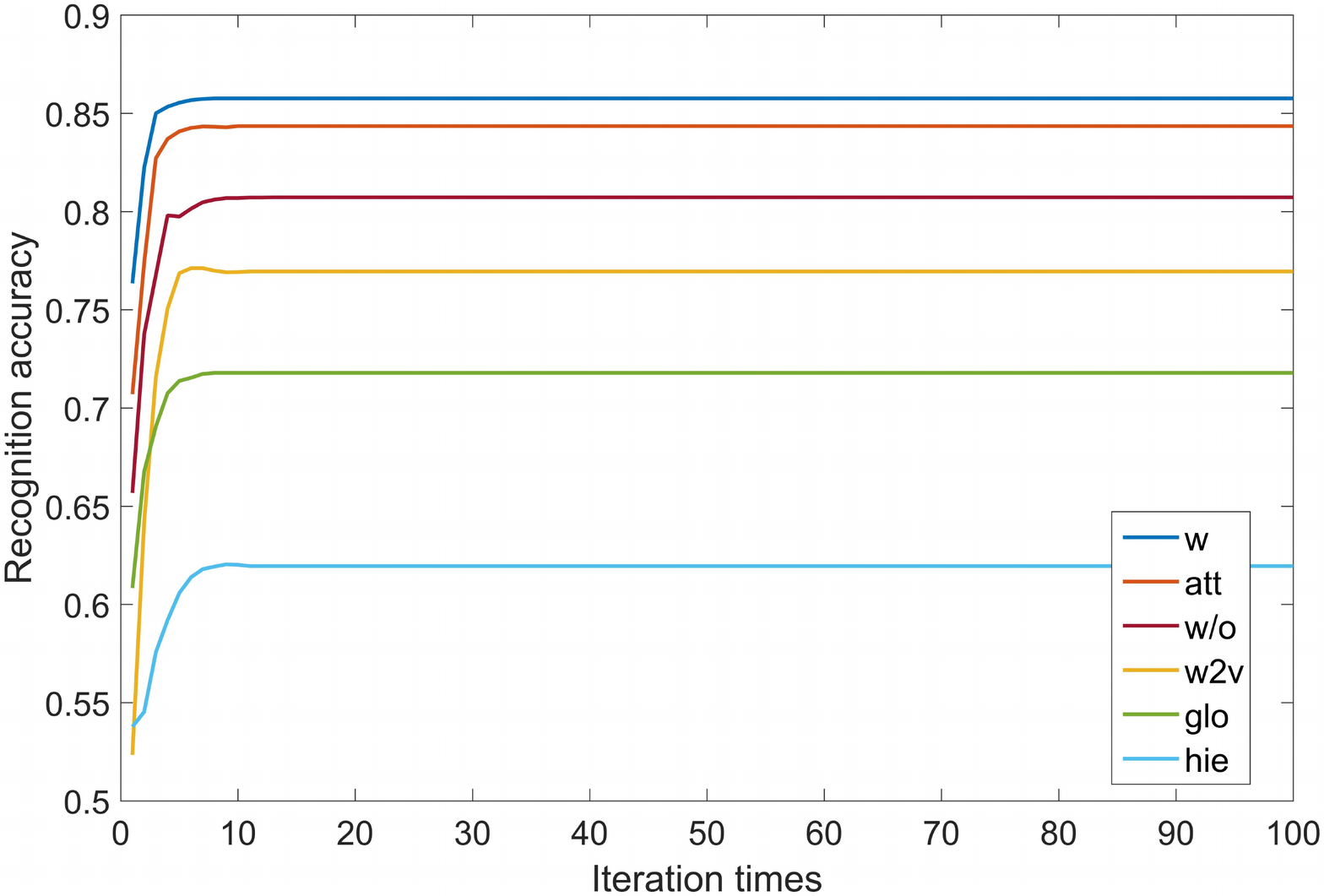}
\end{center}
\vspace{-0.2in}
 \caption{Average per-class Top-1 accuracy(\%) of unseen classes is reported with structure propagation iteration times on AwA. w: the fusion includes att, w2v, glo and hie, while w/o: the fusion contains w2v, glo and hie}
  \label{fig6}
 \end{figure*}

The main idea of the proposed SP method shows three contents. In the first content, the manifold structure of images is considered for constructing the compatibility function between the class label and the visual feature. In the second content, the relationship between multi-manifold structures is found for booting the influence of the positive structure. In the last content, it is the most important to propagate the positive structure and fuse multi-manifold structures by the iteration computation. Therefore, we carry out the related experiment for evaluating the effect of the iteration on the structure evolution in AwA. The recognition accuracy can show the approximation degree of the class manifold structure. In other word, the better recognition accuracy is proportional to the more similar relationship between the reconstruction manifold structure and the intrinsic manifold structure of classes. Figure \ref{fig6} demonstrates the recognition accuracy change with the iteration. In the beginning, the recognition accuracy rapidly increases with the iteration, and then reaches a stable state. It means that structure propagation with the iteration can advance the recognition accuracy and finally obtain the best state.

\subsection{Comparison with state-of-the-arts}
In term of the image data utilization of unseen classes in testing, we can divide ZSL methods into two categories, which are inductive ZSL and transductive ZSL. Inductive ZSL methods can serially process unseen samples without the consideration of the underlying manifold structure in unseen samples\cite{7298911} \cite{7780384} \cite{Changpinyo2016} \cite{zhang2015zero}, while transductive ZSL can usually use the manifold structure of unseen samples to improve ZSL performance \cite{fu2015transductive} \cite{Kodirov2015Unsupervised} \cite{Li2017Paths}. SP can find the structure of unseen classes in image feature space to enhance the transfer model between seen and unseen classes, so SP belongs to a transductive ZSL method. For a fair comparison, we use deep feature of images based on GoogleNet\cite{7298594} in contrasting methods, which include our method, one transductive ZSL method (DMaP \cite{Li2017Paths}), and three inductive ZSL methods(SJE\cite{7298911}, LatEm\cite{7780384} and SynC\cite{Changpinyo2016}). To the best of our knowledge, these methods are state-of-the-art methods for ZSL. Table \ref{table10} shows their results for ZSL on three benchmark datasets. SP mostly outperforms the state-of-the-art methods except DMaP on CUB. DMaP focuses on the manifold structure consistency between the semantic representation and the image feature, and can better distinguish fine-grained classes. SP can complement the manifold structure between the semantic representation and the image feature, and better recognize coarse-grained classes. Therefore,integrating two ideas is expected to further improve the ZSL performance in future work.

\begin{table}[!ht]
\small
\renewcommand{\arraystretch}{1.3}
\caption{Comparison of SP method with state-of-the-art methods for ZSL, average per-class Top-1 accuracy (\%) of unseen classes is reported based on the same data configurations. '+' indicates fusion operation.}
\label{table10}
\begin{center}
\newcommand{\tabincell}[2]{\begin{tabular}{@{}#1@{}}#2\end{tabular}}
\begin{tabular}{lp{2cm}p{1cm}p{1cm}p{1cm}p{1cm}}
\hline
\bfseries Method &\bfseries Semantic feature &\bfseries T/I &\bfseries AwA &\bfseries CUB &\bfseries Dogs\\
\hline \hline
SJE  & att &I& $66.7$ & $50.1$ & N/A\\
     & w2v &I& $51.2$ & $28.4$ & $19.6$\\
\hline
LatEm  & att &I& $71.9$ & $45.5$ & N/A\\
       & w2v &I& $61.1$ & $31.8$ & $22.6$\\
\hline
SynC  & att &I& $69.3$ & $47.5$ & N/A\\
      & w2v &I& $52.9$ & $32.3$ & $27.6$\\
\hline\hline
DMaP & att &T& $74.9$ & $\textbf{61.8}$ & N/A\\
     & w2v &T& $67.9$ & $31.6$ & $38.9$\\
     & att+w2v &T& $78.6$ & $59.6$ & N/A\\
\hline
SP & att &T& $84.3$ & $51.8$ & N/A\\
   & w2v &T& $77.4$ & $32.5$ & $33.3$\\
   & att+w2v &T& $84.7$ & $52.5$ & N/A\\
   & att+w2v+glo+hie &T& $\textbf{85.4}$ & $54.1$ & N/A\\
   & w2v+glo+hie &T& $81.4$ & $35.3$ & $\textbf{48.1}$\\
\hline
\end{tabular}
\end{center}
\end{table}

\subsection{Experimental result analysis}
From the above mention, five methods for constructing the compatibility function have different consideration of the manifold structure. SJE can structure the output space by weighting the different output embedding, which can be associated with the confidence contribution. LatEm try to find the structured model for making the overall piecewise linear function and can capture the flexible model of the latent space for fitting the unseen class. SynC can consider the manifold structure in semantic space for achieving optimal discriminative performance in the model space. DMaP can construct the manifold structure consistency between semantic representation and image feature. The proposed SP can take into consideration for optimizing the relationship of the manifold structure in semantic and image space, and enhance the positive structure propagation by iteration computation for ZSL. From the above experiments, we can attain the following observations.
\begin{itemize}
\item The semantic description have the different contribution for classifying unseen classes. The supervised attribute tend to obtain the better recognition performance than the unsupervised semantic representation (w2v, glo and hie) in AwA and CUB. In the unsupervised semantic representation, the recognition accuracy of w2v or glo is better than that of hie in in AwA and CUB, but the performance of hie is superior to that of  w2v or glo in Dogs.  This is mainly due to that the flexibility and uncertainty of the semantic representation in the unsupervised way.
\item The performance of SP is better than that of other three methods, which are SJE, LatEm,and SynC. However, the performance improvement is different in the various datasets. The obvious improvement can be found in AwA, Dogs and SUN, while the slight improvement can be shown in CUB. The main reason of this situation is related to whether or not effectively to propagate the positive structure in the optimization computation in term of data differences.
\item SP emphasizes on the different manifold structure complement, while DMaP focuses on the various manifold structure consistency. Therefore, the performance of SP is superior to that of DMaP because the structure complementarity plays the important role for learning transfer model in AwA and Dogs, and the performance of DMaP is better than that of SP because the structure consistency is a key point for classifying unseen classes in CUB.
\item SP performs better with the positive structure propagation. SP has demonstrated great promise in above experiments due to multi-manifold structure consideration and alternated optimization between the weight computation and the manifold structure estimation for ZSL.
\item  The proposed fusion method can attain the better performance than the non-fusion method because of appropriate complementing each other. w or w/o always performs better on AwA, CUB and Dogs.
\item The most computational load involved in SP is for solving quadratic programming problem. Specifically,the complexity is $O(PU^{3}+Pk^{3})$ for $P$ iteration times.
\end{itemize}

\section{Conclusion}
We have proposed a new ZSL method, which called Structure Propagation (SP). This method can not only directly model the relevance among the manifold structures in semantic and image space, but also dynamically propagate the positive structure by the crossing iteration. Specifically, the proposed SP method mainly includes four parts. First, nonlinear model constructs the mapping relationship between the class label and the visual image representation. Second, graph describes the relevance between seen classes and unseen classes in semantic or image space. Three, loss function indicates the constrains relationship of multi-manifold structure to balance the structure dependance. Last, structure propagation is implemented by the crossing iteration computation between phantom classes and weights solving. For evaluating the proposed SP, we carry out the experiment on AwA, CUB, Dogs and SUN. Experimental results show that SP can obtain the promising results for ZSL.

\section{Acknowledgements}
The authors would like to thank the anonymous reviewers for their insightful comments that help improve the quality of this paper. Especially, The authors thank to Dr. Yongqin Xian from MPI for Informatics, who provided the data source to us. This work was supported by NSFC (Program No.61771386), Natural Science Basic Research Plan in Shaanxi Province of China (Program No.2016JM6045,No.2017JZ020).

\section*{References}

\bibliography{mybibfile}

\end{document}